\documentclass[11pt]{article}
\usepackage{amssymb}
\usepackage{amsmath, amsthm, color}
\usepackage[colorlinks]{hyperref}
\usepackage{fullpage}

\usepackage[T1]{fontenc}
\usepackage[margin=1.5in]{geometry}    

\usepackage[english]{babel}
\usepackage[utf8x]{inputenc}
\usepackage{algorithm}

\usepackage{caption}
\usepackage{subcaption}

\usepackage{makecell}

\usepackage{bbm}

\usepackage{graphicx}

\usepackage{chngcntr}

\usepackage{mathtools}

\def\sign{{\mathrm{sign}}}

\theoremstyle{definition}
\newtheorem*{notation*}{Notation}
\newtheorem{theorem}{Theorem}

\newtheorem{proposition}{Proposition}

\theoremstyle{definition}
\newtheorem{remark}{Remark}
\newlength{\widebarargwidth}
\newlength{\widebarargheight}
\newlength{\widebarargdepth}

\newcommand{\RNum}[1]{\uppercase\expandafter{\romannumeral #1\relax}}

\begin{document}
\title
{\bf Boosting Algorithms for Estimating Optimal Individualized Treatment Rules}

\makeatletter
\renewcommand\@date{{%
  \vspace{-\baselineskip}%
  \large\centering
  \begin{tabular}{@{}c@{}}
    Duzhe Wang\textsuperscript{$\ast$} \\
    \normalsize \texttt{dwang282@wisc.edu}
  \end{tabular}%
  \quad \quad
  \begin{tabular}{@{}c@{}}
    Haoda Fu\textsuperscript{$\dagger$} \\
    \normalsize  \texttt{fu$\_$haoda@lilly.com}
     \end{tabular}
 \quad \quad
  \begin{tabular}{@{}c@{}}
    Po-Ling Loh\textsuperscript{$\ast\ddagger$} \\
    \normalsize  \texttt{ploh@stat.wisc.edu}
     \end{tabular}

  \bigskip

  \textsuperscript{$\ast$}Department of Statistics, University of Wisconsin-Madison\par
  \textsuperscript{$\dagger$}Advanced Analytics and Data Sciences, Eli Lilly and Company \par
  \textsuperscript{$\ddagger$}Department of Statistics, Columbia University
    \bigskip

  \today
}}
\makeatother

\maketitle

\begin{abstract}
We present nonparametric algorithms for estimating optimal individualized treatment rules. The proposed algorithms are based on the XGBoost algorithm, which is known as one of the most powerful algorithms in the machine learning literature. Our main idea is to model the conditional mean of clinical outcome or the decision rule via additive regression trees, and use the boosting technique to estimate each single tree iteratively. Our approaches overcome the challenge of correct model specification, which is required in current parametric methods. The major contribution of our proposed algorithms is providing efficient and accurate estimation of the highly nonlinear and complex optimal individualized treatment rules that often arise in practice. Finally, we illustrate the superior performance of our algorithms by extensive simulation studies and conclude with an application to the real data from a diabetes Phase III trial. 
\end{abstract}

%%%%%%%%%%%%%%%%%%%%%%%%%%%%%%%%%%%%%%%%%%%%%%%%%%%%%%%%%%%%
\section{Introduction}

Precision medicine, as an emerging medical approach for disease treatment and prevention, has received more and more attention among government, healthcare industry and academia in recent years. It is a well-known fact that there exists a significant heterogeneity for patients in response to treatments. For example, as demonstrated in \cite{havlir2011}, for patients who are infected with human immunodeficiency virus and tuberculosis, their optimal timing of antiretroviral therapy (ART) varies significantly. \cite{havlir2011} concludes that patients with CD4+ T-cell counts of less than 50 per cubic millimeter receive substantial benefits from the earlier ART, while those with larger CD4+ T-cell counts don't.  Therefore, compared with the traditional one-size-fits-all approach, precision medicine aims to optimize clinical outcome by taking into account individual variability in clinical features, genes, environment, behaviors, and habits for each person.  

In drug development, there are usually multiple available treatments for the same disease. For instance, there are different classes of oral and injectable treatments for treating Type II diabetes mellitus \cite{fu2014}. This motivates one current active line of research in statistics and machine learning, which is called individualized treatment recommendation. Generally speaking, the goal of individualized treatment recommendation is to decide an optimal rule which assigns a treatment from the set of possible ones to a patient based on their clinical characteristics. During the last decade, there is a large amount of literature on estimating the optimal individualized treatment rule \cite{murphy2003optimal, qian2011, zhao2012, liu2016robust, zhou2017, chen2017, qi2019}. In general, there are two frameworks, namely indirect learning and direct learning, among existing approaches to estimate the optimal individualized treatment rules. The key idea of indirect learning is first estimating the conditional mean outcome and then determining the optimal treatment rule by comparing the conditional means across various treatments \cite{qian2011}. In contrast, approaches in direct learning estimate the decision rule directly via optimizing some objective functions. For example, \cite{zhao2012} proposes the outcome weighted learning (OWL), which transforms the value maximization problem to a weighted 0-1 loss minimization problem. Then OWL replaces the 0-1 loss by the hinge loss and uses techniques in the literature of support vector machine to seek the optimal rule. \cite{liu2016robust} and \cite{zhou2017} further improve the performance of OWL by carefully choosing the weights. Another example of direct learning is  \cite{qi2019}, which represents the optimal decision rule as a minimizer of a weighted least squares problem and proposes to optimize the corresponding empirical risk. 

Indirect learning and direct learning open the door for statisticians and machine learning researchers to bring data-driven approaches to the area of precision medicine. 
However, there is still significant room for developing new efficient and accurate methods. First, most of the existing approaches in both of the indirect learning and direct learning assume a parametric linear form either for the conditional expectation of the outcome variable or for the decision rule. Therefore, the success of these approaches highly depends on the correct specification of the posited models. However, it is a very common opinion that decision rules which characterize the relationship between the clinical variables and the clinical response are highly nonlinear. Second, current proposed algorithms to solve objectives in indirect learning and direct learning don't scale for large-scale datasets. Third, the original OWL and most of its variants do not include the variable selection procedure, which is of great importance to improve the estimation accuracy when there is a large amount of clinical covariates. These challenges may result in bad performance of existing approaches in practice.  

To alleviate problems discussed above, we develop several new methods, which model the conditional expectation of outcome variable and the decision rule by additive trees. Our proposed approaches use XGBoost-based boosting algorithms to estimate each regression tree iteratively. As a powerful boosting algorithm developed recently in \cite{chen2016xgboost}, XGBoost has been widely applied in different kinds of regression and classification problems in the machine learning community. The main contributions of our work are summarized as follows:
\begin{itemize}
\item Propose the flexible nonparametric framework which uses additive trees to model the conditional expectation and decision rule. This not only enlarges the model space, but also improves the estimation accuracy compared with methods using a single tree. 
\item Present novel XGBoost-based boosting algorithms for both of the indirect learning and direct learning. To the best of our knowledge, this work is the first study to apply the XGBoost algorithm to the area of estimating optimal individualized treatment rules. 
\item Develop a new direct learning approach (i.e., objective (\ref{owl3}) in Section~\ref{bstdir}) and provide a theoretical analysis on Fisher consistency of our proposed method. 

\item Demonstrate that our algorithms outperform other existing methods in a wide variety of settings via extensive simulation studies and an application to the real data from a diabetes Phase III trial.

\end{itemize}

The remainder of our paper is organized as follows: Section~\ref{backgroundsec} furnishes the mathematical
background for the individualized treatment rule, indirect learning and direct learning to be considered in the paper. Section~\ref{methodsec} presents our main methodological and theoretical contributions, providing the tree boosting algorithms and deriving the Fisher consistency of our proposed method in direct learning. Section~\ref{simusec} contains extensive simulation studies that are used to validate the performance of our proposed methods. In Section~\ref{realdatasec}, we apply our methods to a clinical trial Phase III diabetes dataset.  
We conclude with a discussion in Section~\ref{dis-sec}, including some avenues for future research.

%%%%%%%%%%%%%%%%%%%%%%%%%%%%%%%%%%%%%%%%%%%%%%%%%%%%

\section{Background and problem setup}
\label{backgroundsec}
In this section, we describe the individualized treatment rules, which are to be studied in our paper. We also discuss several popular existing approaches in the literature. 

\subsection{Individualized treatment rules}

We observe a data set $(X_{i}, A_{i}, Y_{i})$ for $1\le i\le n$, where $X_{i}\in \mathcal{X}\subset\mathbb{R}^{p}$ is the $i$-th patient's prognostic variable, $A_{i}\in \mathcal{A}$ is the $i$-th patient's treatment assignment, and $Y_{i}\in \mathbb{R}$ is the $i$-th patient's clinical outcome. In this paper, we focus on the binary treatments. That is, we assume $\mathcal{A}=\{-1, +1\}$. Extensions to the multinary case where there are more than two treatments will be discussed in Section~\ref{dis-sec}. 

Furthermore, we assume $\left(X_{i}, A_{i}, Y_{i}\right)$ are i.i.d.\ copies of $(X, A, Y)$, which has an unknown underlying distribution $P$. The clinical outcome variable $Y$ is also called reward in the literature of individualized treatment rules. In general, we assume that higher values of $Y$ are better. Let $f_{0}(x)$ be the density function of $X$. For $a\in \mathcal{A}$, let $\pi_{a}(x)=P(A=a|x)$ be the conditional probability of $A=a$ given $X=x$. In this paper, we focus on the randomized controlled trial where $A$ is independent of $X$ and $\pi_{a}(x)$ is \emph{known}. Let $f_{1}(y|X=x, A=a)$ be the conditional density of $Y$ given $X=x$ and $A=a$. Then the joint probability density function of $(X,  A, Y)$ can be written as 
\begin{equation*}
f(x, a, y)=f_{0}(x)\pi_{a}(x)f_{1}(y|x, a). 
\end{equation*}

An individualized treatment rule (ITR) $\mathcal{D}$ is a deterministic function, which maps from the covariate space $\mathcal{X}$ to the treatment set $\mathcal{A}$. For a fixed ITR $\mathcal{D}$, let $A^{\mathcal{D}}=\mathcal{D}(X)$. We assume $(X, A^{\mathcal{D}}, Y)$ have a joint distribution, denoted by $P^{\mathcal{D}}$, and we denote the corresponding 
joint density function by $f^{\mathcal{D}}(x, a, y)$. Then we have
\begin{equation*}
f^{\mathcal{D}}(x, a, y)=f_{0}(x)I(a=\mathcal{D}(x))f_{1}(y|x, a). 
\end{equation*}
Let $E^{\mathcal{D}}$ be the expectation operator with respect to $P^{\mathcal{D}}$. Then we define the following \emph{value function} associated with an ITR $\mathcal{D}$, which is used to describe the average clinical outcome when assigning treatments via the ITR $\mathcal{D}$: 
\begin{equation*}
 V(\mathcal{D})=E^{\mathcal{D}}(Y)=\int YdP^{\mathcal{D}}. 
\end{equation*}
Furthermore, it is straightforward to show that 
\begin{equation}
\label{valuefunction}
V(\mathcal{D})=\int Y\frac{dP^{\mathcal{D}}}{dP}dP=E\left\{Y\frac{I(A=\mathcal{D}(X))}{\pi_{A}(X)}\right\}, 
\end{equation}
where the expectation operator $E$ is with respect to the underlying joint distribution $P$. 

An optimal individualized treatment rule, denoted by $\mathcal{D}^*$, is the treatment assignment which maximizes the value function $(\ref{valuefunction})$. That is, 
\begin{equation}
\label{optitr}
\begin{aligned}
&\mathcal{D}^*=\underset{\mathcal{D}}{\text{argmax}}
& & V(\mathcal{D}). 
\end{aligned}
\end{equation}
In this paper, we are interested in estimating the optimal ITR $\mathcal{D}^*$ using observed data $\{ (X_{i}, A_{i}, Y_{i}), 1\le i\le n  \}$. 

%%%%%%%%%%%%%%%%%%%%%%%%%%%%%%%%%%%%%%%%%%%%%%%%%%%%%%%%%%
\subsection{Previous work}
\label{previouswork}

We now briefly introduce two different lines of previously proposed work for estimating the optimal ITR for binary treatments. We start with an indirect learning approach, which is often called Q-learning in the machine learning literature. The following Proposition provides an equivalent form of the optimal ITR defined in $(\ref{optitr})$. 
\begin{proposition}
\label{equivalent-itr}
Let $Q(X, A)=E(Y|X, A)$. Then the optimal ITR defined in $(\ref{optitr})$ satisfies 
\begin{equation*}
\begin{aligned}
&\mathcal{D}^*(X)= \underset{a\in \mathcal{A}}{\text{argmax}}
& & Q(X, a)\quad\mathrm{a.s.}
\end{aligned}
\end{equation*}
\end{proposition}
The proof of Proposition~\ref{equivalent-itr} is contained in Appendix~\ref{equivalent-itr-proof}. The conditional expectation $Q(x, a)$ is usually called the \emph{quality} of treatment $a$ at observation $x$. Therefore, one standard regression-based approach to estimate the optimal ITR from Proposition~\ref{equivalent-itr} uses parametric forms of $Q(x, +1)$ and $Q(x, -1)$. For example,  current Q-learning assumes 
\begin{equation*}
Q\left(x, +1\right)=\beta_{0}^{1}+\beta_{1}^Tx, \quad Q\left(x, -1\right)=\beta_{0}^{-1}+\beta_{-1}^Tx,
\end{equation*}
 and estimates two sets of regression coefficients $(\beta_0^{1}, \beta_{1}^T)$ and $(\beta_{0}^{-1}, \beta_{-1}^T)$ respectively by either the ordinary least squares or penalized least squares. For example, the ordinary least squares approach solves 
 \begin{equation}
 \label{indir1}
\begin{aligned}
& \left(\hat{\beta}_{0}^{1}, \hat{\beta}_{1}^T\right)= \underset{\beta_{0}^1,~\beta_{1}}{\text{argmin}}
& & \sum_{i:A_i=1}\left(Y_i-\beta_{0}^1-\beta_{1}^TX_i\right)^2,
\end{aligned}
\end{equation} 
and 
 \begin{equation}
 \label{indir2}
\begin{aligned}
& \left(\hat{\beta}_{0}^{-1}, \hat{\beta}_{-1}^T\right)= \underset{\beta_{0}^{-1},~\beta_{-1}}{\text{argmin}}
& & \sum_{i:A_i=-1}\left(Y_i-\beta_{0}^{-1}-\beta_{-1}^TX_i\right)^2. 
\end{aligned}
\end{equation}
 In the end, Q-learning indirectly estimates $\mathcal{D}^*(x)$ via the sign function of difference of the estimated $Q(x, +1)$ and $Q(x, -1)$.  That is, 
 \begin{equation}
 \label{indirectlearning}
 \widehat{\mathcal{D}}\left(x\right)=\text{sign}\left(\hat{\beta}_{0}^{1}-\hat{\beta}_{0}^{-1}+\left(\hat{\beta}_{1}^{T}-\hat{\beta}_{-1}^{T}\right)x\right). 
 \end{equation}
 Equation $(\ref{indirectlearning})$ implies that the decision rule is linear with respect to covariates $x$. 

Rather than estimating $Q(x, +1)$ and $Q(x, -1)$ separately, \cite{qian2011} considers a slightly different approach, which models $Q(X, A)$ via basis functions from the interaction space of clinical covariates and treatments $\mathcal{X}\times \mathcal{A}$. More specifically, \cite{qian2011} approximates $Q(X, A)$ via 
\begin{equation*}
Q(X, A)=(1, X, A, XA)\theta,
\end{equation*}
where $\theta$ is a vector in $\mathbb{R}^{2p+2}$. Then \cite{qian2011} solves the following objective
 \begin{equation}
 \label{indir3}
\begin{aligned}
& \underset{\theta}{\text{minimize}}
& & \sum_{i=1}^{n}\left\{Y_i-(1, X_i, A_i, X_iA_i)\theta\right\}^2+\lambda\|\theta\|_{1}. 
\end{aligned}
\end{equation} 

Next, we introduce two direct learning approaches for estimating the optimal ITR. Let
\begin{equation*}
f^*(x)=Q(x, +1)-Q(x, -1)=E\left\{\frac{YA}{\pi_{A}(X)}|X=x\right\}. 
\end{equation*}
Then Proposition~\ref{equivalent-itr} implies that 
\begin{equation}
\label{signdecision}
\mathcal{D}^*\left(x\right)=\text{sign}(f^*(x))=
 \begin{cases} 
      -1 &  f^*(x)<0\\
     +1 &  f^*(x)>0. 
   \end{cases}
\end{equation}
Therefore, we can directly estimate the decision rule $f^*(x)$ and then use the sign of the estimated function. Furthermore, we have the following proposition: 
\begin{proposition}
\label{estflemma}
Under the change of differential and expectation condition, $f^*$ is an optimal solution to 
\begin{equation*}
\begin{aligned}
& \underset{g}{\text{argmin}}
& & E\left\{\frac{1}{\pi_{A}(X)}(2YA-g(X))^2\right\}. 
\end{aligned}
\end{equation*}
\end{proposition} 
Proposition~\ref{estflemma} is similar to  results in \cite{qi2019} and its proof is contained in Appendix~\ref{estflemmaproof}. If we assume $f^*(x)$ is linear, that is, 
$f^*(x)=\beta^*_0+(\beta^*)^Tx$, then Proposition~\ref{estflemma} suggests that we can estimate $\beta_0^*$ and $\beta^*$ via the empirical risk minimization
\begin{equation}
\label{opt1}
\left(\hat{\beta}_{0}, \hat{\beta}\right)=
\begin{aligned}
& \underset{\beta_0,~\beta}{\text{argmin}}
& & \sum_{i=1}^{n}\frac{1}{\pi_{A_i}(X_i)}(2Y_iA_i-\beta_0-\beta^TX_i)^2
\end{aligned}
\end{equation}
when the number of covariates is small, or via the regularized risk minimization 
\begin{equation}
\label{opt2}
\left(\hat{\beta}_{0}, \hat{\beta}\right)=
\begin{aligned}
& \underset{\beta_0,~\beta}{\text{argmin}}
& & \sum_{i=1}^{n}\frac{1}{\pi_{A_i}(X_i)}(2Y_iA_i-\beta_0-\beta^TX_i)^2+\lambda\|\beta\|_{1}
\end{aligned}
\end{equation} 
in the high-dimensional setting. 

Another popular work in the line of direct learning is called outcome weighted learning \cite{zhao2012}.  For a fixed ITR $\mathcal{D}$, it's easy to see that 
\begin{equation*}
V(\mathcal{D})+E\left\{Y\frac{I(A\ne \mathcal{D}(X))}{\pi_{A}(X)}\right\}=E\left\{\frac{Y}{\pi_{A}(X)}\right\}, 
\end{equation*}
which is a constant. Therefore, the value maximization problem $(\ref{optitr})$ is equivalent to the following risk minimization problem
\begin{equation*}
\mathcal{D}^{*}=
\begin{aligned}
& \underset{\mathcal{D}}{\text{argmin}}
& &  E\left\{Y\frac{I(A\ne \mathcal{D}(X))}{\pi_{A}(X)}\right\}.
\end{aligned}
\end{equation*}
Hence by equation $(\ref{signdecision})$, we have
\begin{equation}
\label{owl}
f^*=
\begin{aligned}
& \underset{f}{\text{argmin}}
& & E\left\{Y\frac{I(Af(X)<0)}{\pi_{A}(X)}\right\},
\end{aligned}
\end{equation}
where $Af(X)$ is usually called the \emph{functional margin}. The risk function in problem $(\ref{owl})$ can be viewed as a weighted expectation of 0-1 loss. It is well known that dealing with 0-1 loss is difficult due to its non-convexity. Therefore, under the assumption that $Y_{i}\ge 0$, for $1\le i\le n$, \cite{zhao2012} replaces the 0-1 loss by the hinge loss and aims to optimize the following convex objective
\begin{equation}
\label{owl2}
\begin{aligned}
& \underset{f\in \mathcal{F}}{\text{minimize}}
& & \frac{1}{n}\sum_{i=1}^{n}\frac{Y_{i}}{\pi_{A_{i}}(X_{i})}(1-A_{i}f(X_{i}))^{+}+\lambda\|f\|^2,
\end{aligned}
\end{equation}
where $u^{+}=\max(u, 0)$, $\|f\|$ is some norm of $f$, and $\mathcal{F}$ is some specified functional space. More specifically, \cite{zhao2012} considers the linear decision rule for optimal ITR, which solves 
\begin{equation}
\label{opt3}
\left(\hat{\beta}_{0}, \hat{\beta}\right)=
\begin{aligned}
& \underset{\beta_0, ~\beta}{\text{argmin}}
& & \frac{1}{n}\sum_{i=1}^{n}\frac{Y_{i}}{\pi_{A_{i}}(X_{i})}(1-A_{i}(\beta_0+\beta^TX_i))^{+}+\lambda\|\beta\|^2. 
\end{aligned}
\end{equation}
When there exist negative outcomes, \cite{zhao2012} subtracts the minimum observed outcome from all outcome responses.

%%%%%%%%%%%%%%%%%%%%%%%%%%%%%%%%%%%%%%%%%%%%%%%%%%%%%%%%%%
\section{Tree boosting algorithms}
\label{methodsec}

While those previous parametric approaches in Section~\ref{previouswork} are easy to interpret, they face several challenges to be addressed. We highlight three key issues here: (1) They may suffer from the issue of model misspecification. On one hand, the linear assumption of $Q(x, +1)$ and $Q(x, -1)$ implicitly lead to the linear decision rule. On the other hand, objectives $(\ref{opt1})$, $(\ref{opt2})$, and $(\ref{opt3})$ directly assume that the optimal decision function $f^*(x)$ is linear with respect to $x$. However,  in practice, the decision rule is usually highly nonlinear. (2) Numerical algorithms which are proposed to solve $(\ref{opt1})$, $(\ref{opt2})$, and $(\ref{opt3})$ are not scalable. Therefore, it is computationally expensive to deal with large-scale data sets. (3) In practice, a large amount of clinical covariates are often available for estimating the optimal ITR, but many of them might not be related to the prediction of outcomes. Hence, optimization problem $(\ref{opt3})$ in outcome weighted learning, which does not incorporate the procedure of variable selection, may have a bad performance. Motivated by these challenges, in this section, we propose several efficient tree boosting algorithms, which are based on the well-known XGBoost algorithm \cite{chen2016xgboost}, for estimating the optimal ITR. 

\subsection{A tree boosting algorithm in indirect learning}
We consider modeling quality functions in the indirect learning by additive trees. That is, we assume 
\begin{equation}
\label{qualitytree1}
Q(x, +1)=\sum_{k=1}^{K}b_{k}^{1}(x), 
\end{equation}
and 
\begin{equation}
\label{qualitytree2}
Q(x, -1)=\sum_{k=1}^{K}b_{k}^{-1}(x),
\end{equation}
where $b_{k}^{1}$ and $b_{k}^{-1}$ are in the space of regression trees defined by 
\begin{equation*}
\mathcal{F}=\{ b(x)=w_{q(x)};\quad q: \mathcal{X}\rightarrow T  \}. 
\end{equation*}
Here, the map $q$ denotes the structure of a tree, $T$ is the set of leaves of a tree, and $w_{q(x)}$ is the outcome of leaf indexed by $q(x)$. Number of regression trees $K$ is \emph{user-defined} in practice. $(\ref{qualitytree1})$ and $(\ref{qualitytree2})$ are examples of boosting regression trees, which is known as one of the most powerful methods in statistical learning. We aim to estimate these basis functions via the following two objectives: 
\begin{equation}
\label{bst1}
\begin{aligned}
& \underset{b_{k}^{1}, ~1\le k\le K}{\text{minimize}}
& & \sum_{i: A_i=1}\left\{Y_i-\sum_{k=1}^{K}b_{k}^{1}(X_i)\right\}^2+\sum_{k=1}^{K}J(b_{k}^{1}), 
\end{aligned}
\end{equation}
and 
\begin{equation}
\label{bst2}
\begin{aligned}
& \underset{b_{k}^{-1}, ~1\le k\le K}{\text{minimize}}
& & \sum_{i: A_i=-1}\left\{Y_i-\sum_{k=1}^{K}b_{k}^{-1}(X_i)\right\}^2+\sum_{k=1}^{K}J(b_{k}^{-1}), 
\end{aligned}
\end{equation}
where $J(b_{k}^{1})$ and $J(b_{k}^{-1})$ are penalties used to control the complexity of regression trees. Following \cite{chen2016xgboost}, in our proposed algorithm,  for a tree $b$, we set 
\begin{equation*}
\label{treepenalty}
J(b)=\gamma |T|+\frac{1}{2}\lambda\|w\|_{2}^2,
\end{equation*}
where $|T|$ is the number of leaves, and $\gamma>0$ and $\lambda>0$ are tuning parameters. In practice, \cite{chen2016xgboost} sets a maximum depth of the regression tree in order to choose the appropriate tuning parameters.

Next, we propose using forward stagewise additive algorithms to solve the above two optimization problems. We take objective $(\ref{bst1})$ as an example. Let $\hat{f}_{t-1}^{1}$ be estimated additive trees at the (t-1)-th iteration. Then at the t-th iteration, we solve 
\begin{equation}
\label{bst3}
\begin{aligned}
& \underset{b}{\text{minimize}}
& & \sum_{i: A_i=1}\left[Y_i-\left\{\hat{f}_{t-1}^{1}\left(X_i\right)+b(X_i)\right\}\right]^2+J(b). 
\end{aligned}
\end{equation}
Minimization problem $(\ref{bst3})$ is a functional optimization, which needs to estimate tree structure and outcome of leaves. For a fixed tree with $I_{j}=\{i: A_{i}=1, q(X_i)=j \}$ as the instance set of leaf $j$, $(\ref{bst3})$ is equivalent to the following minimization problem with respect to $w$: 
\begin{equation}
\label{bst4}
\begin{aligned}
& \underset{w}{\text{minimize}}
& & -2\sum_{j=1}^{|T|}\left[\sum_{i\in I_{j}}\left\{Y_i-\hat{f}_{t-1}^{1}\left(X_i\right)\right\}\right]w_{j}+\sum_{j=1}^{|T|}\left(\frac{1}{2}\lambda+|I_j|\right)w_{j}^2+\gamma |T|. 
\end{aligned}
\end{equation}
It is straightforward to show that the optimal solution to $(\ref{bst4})$ is 
\begin{equation}
\label{weights}
\hat{w}_{j}=\frac{2\sum_{i\in I_{j}}\left\{Y_i-\hat{f}_{t-1}^{1}\left(X_i\right)\right\}}{\lambda+2|I_j|},
\end{equation}
and the corresponding optimal objective value is 
\begin{equation*}
-2\sum_{j=1}^{|T|}\frac{\left[\sum_{i\in I_{j}}\left\{Y_i-\hat{f}_{t-1}^{1}\left(X_i\right)\right\}\right]^2}{\lambda+2|I_j|}+\gamma |T|. 
\end{equation*}
This optimal value can be regarded as a score function to measure the quality of a tree. It remains to estimate the tree structure for problem $(\ref{bst3})$. Since it's impossible to enumerate all possible tree structures, we apply greedy split finding algorithms proposed in \cite{chen2016xgboost}. The key idea is starting from a single leaf and iteratively adding branches to the tree. For one node with instance set $I$, let $I_{l}$ and $I_{r}$ be instance sets of left and right nodes after the split. Note that we have $I=I_{l}\cup I_{r}$. Then the loss reduction after the split is 
\begin{equation*}
2\left[\frac{\left\{\sum_{i\in I_{l}}\left(Y_i-\hat{f}_{t-1}^{1}\left(X_i\right)\right)\right\}^2}{\lambda+2|I_{l}|}+\frac{\left\{\sum_{i\in I_{r}}\left(Y_i-\hat{f}_{t-1}^{1}\left(X_i\right)\right)\right\}^2}{\lambda+2|I_{r}|}-\frac{\left\{\sum_{i\in I}\left(Y_i-\hat{f}_{t-1}^{1}\left(X_i\right)\right)\right\}^2}{\lambda+2|I|}\right]-\gamma. 
\end{equation*}
This is the main criterion to evaluate the split candidates. For more details of split finding algorithms, we refer the interested reader to Section 3 of \cite{chen2016xgboost} and the references therein. Let $\hat{b}$ be the estimated tree by the greedy split finding algorithms with outcomes of leaves decided by $(\ref{weights})$. Finally, to avoid overfitting, the estimated additive trees at the t-th iteration is given by 
\begin{equation}
\label{shrinkage}
\hat{f}_{t}^{1}=\hat{f}_{t-1}^{1}+\eta\hat{b}, 
\end{equation}
where $0<\eta<1$ is a shrinkage parameter and it needs to be tuned in practice. 

The above algorithm is implemented in the open-source software library of the XGBoost algorithm. To sum up, we propose the following tree boosting algorithm: 
\begin{algorithm}[H]
%\hline
{\bf Input}: data set $\{(X_{i}, A_{i}, Y_{i})\}_{i=1}^{n}$, number of iterations $K$, shrinkage parameter $\eta$ and maximum tree depth $d$. 
 \begin{enumerate}
    \item For observations with $A_i=+1$, use the XGBoost algorithm to optimize $(\ref{bst1})$. Following the same notation used in this section, we denote the estimator of $Q(x, +1)$ by $\hat{f}_{K}^{1}(x)$. 
    \item For observations with $A_i=-1$, use the XGBoost algorithm to optimize $(\ref{bst2})$, and denote the estimator of $Q(x, -1)$ by $\hat{f}_{K}^{-1}(x)$. 
    \item Output the estimated optimal ITR: 
    \begin{equation*}
    \widehat{\mathcal{D}}\left(x\right)=\sign\left(\hat{f}_{K}^{1}\left(x\right)-\hat{f}_{K}^{-1}\left(x\right)\right). 
    \end{equation*}
 \end{enumerate}
 \caption{Tree boosting algorithm in indirect learning} 
 \label{alg1}
\end{algorithm}

\begin{remark}
We provide a caveat while using Algorithm~\ref{alg1}. This algorithm divides the whole data set into two classes: one with treatment $+1$ and the other with treatment $-1$. When the number of observations belonging to one class is significantly lower than those belonging to the other classes, this algorithm may come across the issue of imbalanced classes, which makes the estimation of one quality function not as efficient as the other one.  
\end{remark}

%%%%%%%%%%%%%%%%%%%%%%%%%%%%%%%%%%%%%%%%%%%%%%%%%%%%%%%%%%%%%%%%%%

\subsection{Tree boosting algorithms in direct learning}
\label{bstdir}
We now consider the direct learning framework. Our first tree boosting algorithm in this section is based on Proposition~\ref{estflemma}.  We assume that the decision rule $f^*(x)$ is additive trees: 
\begin{equation*}
f^{*}(x)=\sum_{k=1}^{K}b_{k}(x). 
\end{equation*}
Then we have the following objective
\begin{equation}
\label{bst6}
\begin{aligned}
& \underset{b_{k}, ~1\le k\le K}{\text{minimize}}
& & \sum_{i=1}^{n}\frac{1}{\pi_{A_i}(X_i)}\left(2Y_iA_i-\sum_{k=1}^{K}b_{k}\left(X_i\right)\right)^2+\sum_{k=1}^{K}J(b_{k}). 
\end{aligned}
\end{equation}
The optimization problem $(\ref{bst6})$ is a weighted least squares problem with weights $1/(\pi_{A_i}(X_i))$ and  responses $2Y_{i}A_{i}$. Therefore, similar to Algorithm~\ref{alg1}, we propose the following algorithm: 
\begin{algorithm}[H]
%\SetAlgoLined
{\bf Input}: data set $\{(X_{i}, A_{i}, Y_{i})\}_{i=1}^{n}$, number of iterations $K$, shrinkage parameter $\eta$ and maximum tree depth $d$. 
 \begin{enumerate}
  %  \item Set $\tilde{Y}_{i}=2Y_iA_i$. 
    \item Use the XGBoost algorithm with the weighted quadratic loss function in $(\ref{bst6})$ to estimate $f^{*}(x)$. We denote the estimator by $\hat{f}(x)$. 
    \item Output the estimated optimal ITR: $\widehat{\mathcal{D}}\left(x\right)=\sign\left(\hat{f}\left(x\right)\right)$. 
 \end{enumerate}
 \caption{Tree boosting algorithm I in direct learning} 
 \label{alg2}
\end{algorithm}

\begin{figure}[h!]
\centering
  \includegraphics[scale=0.6]{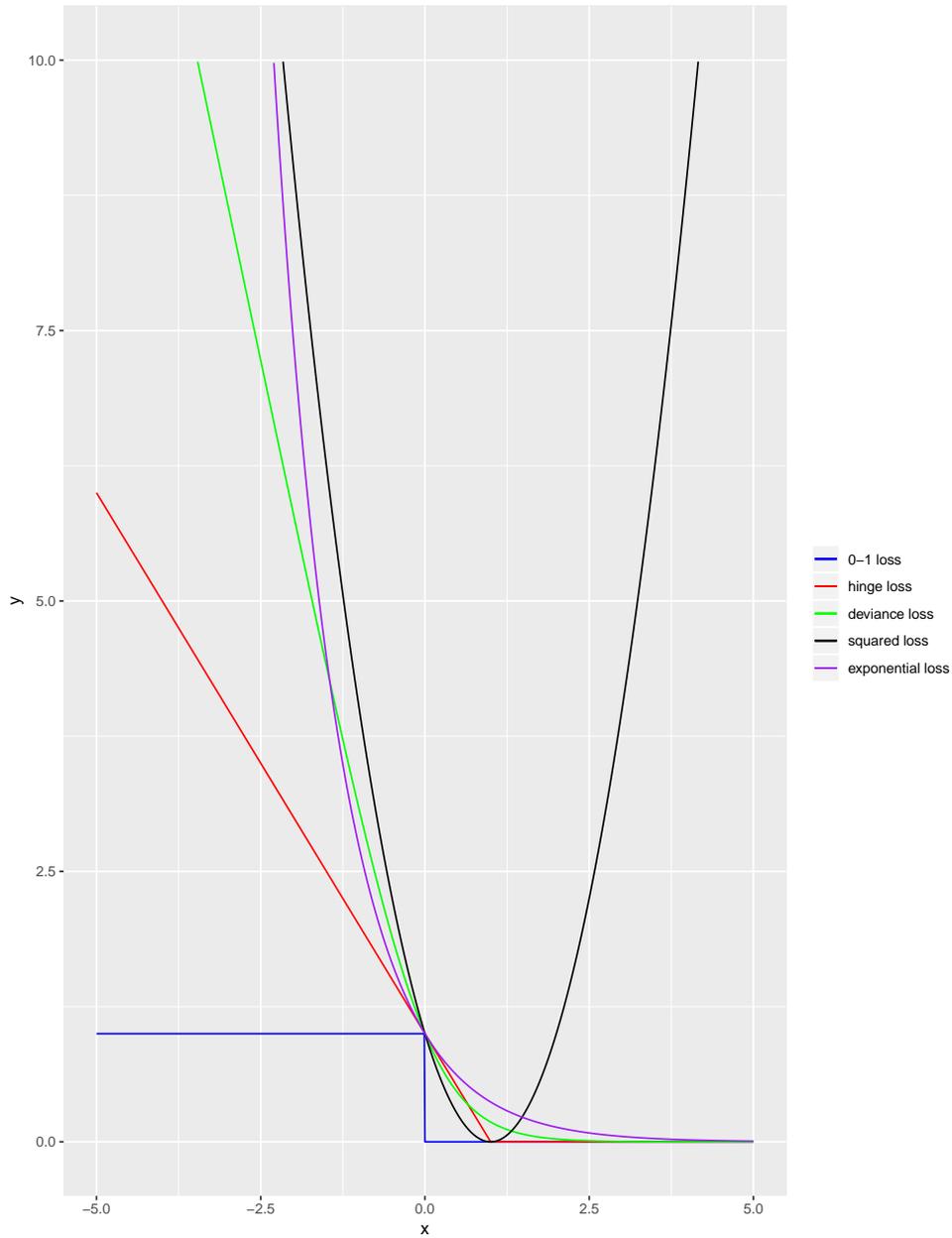}
  \caption{0-1 loss: $I(x<0)$; hinge loss: $(1-x)^{+}$; deviance loss: $\log_{2}(1+e^{-2x})$; squared loss: $(1-x)^2$; exponential loss: $e^{-x}$. Note that the deviance loss in this figure uses base 2, but for simplicity, we use base $e$ in our paper, which is up to a constant factor. }
  \label{lossfunc}
\end{figure}

Next, we consider the outcome weighted learning. Let $\phi$ be some margin-based convex loss function and define 
\begin{equation}
\label{owlconvex}
f^*_{\phi}=
\begin{aligned}
& \underset{f}{\text{argmin}}
& &  E\left\{Y\frac{\phi(Af(X))}{\pi_{A}(X)}\right\}. 
\end{aligned}
\end{equation}
Then we have the following result concerning Fisher consistency of $(\ref{owlconvex})$ . 
\begin{theorem}
\label{fisherconsistency-binary}
Assume $Y$ is nonnegative. For a convex loss function $\phi(x)$, if $\phi(x)$ is differentiable at 0 and $\phi^{\prime}(0)<0$, then for $x\in \mathcal{X}$, we have 
\begin{equation*}
\mathcal{D}^*(x)=\text{sign}(f^*_{\phi}(x)). 
\end{equation*}
\end{theorem}
The proof of Theorem~\ref{fisherconsistency-binary} is contained in Appendix~\ref{fisherconsistency-binary-proof}. In this paper we focus on boosting algorithms based on XGBoost, which requires that the loss function $\phi$ has the second derivative. So we consider the following deviance loss 
\begin{equation*}
\phi(x)=\log (1+e^{-2x}). 
\end{equation*}
Note that squared loss and exponential loss also satisfy conditions in Theorem~\ref{fisherconsistency-binary} and have second derivatives, but deviance loss is known to be more robust \cite{hastie2005elements}. Figure~\ref{lossfunc} shows the comparison of 0-1 loss, hinge loss, deviance loss, squared loss, and exponential loss. 

Furthermore, it is known that results from the outcome weighted learning in $(\ref{owl2})$ are not stable due to the large variability of weights \cite{fu2014, liu2016robust, zhou2017}. Therefore, to further improve the finite sample performance, we apply similar ideas in \cite{liu2016robust} and \cite{zhou2017}. We assume that the outcome variable $Y$ has the representation
\begin{equation}
\label{responserep}
Y=\mu(X)+\delta(X)\times A+\epsilon,
\end{equation}
where $\epsilon$ is the random error with zero mean, $\mu(X)$ is the common effect of covariates $X$, and $\delta(X)\times A$ is the interaction effect between covariates and the treatment. Then we have the following result: 
\begin{theorem}
\label{theorem2}
Let 
\begin{equation}
\label{minimizer}
f^{**}_{\phi}=
\begin{aligned}
& \underset{f}{\text{argmin}}
& & E\left\{\frac{|Y-\mu(X)|}{\pi_{A}(X)}\phi\left(Af(X)\times \text{sign}(Y-\mu(X))\right)\right\},
\end{aligned}
\end{equation}
where $\phi(x)=\log(1+e^{-2x})$. Then we have 
\begin{equation*}
\mathcal{D}^*(x)=\text{sign}(f^{**}_{\phi}(x)). 
\end{equation*}
\end{theorem}
The proof of Theorem~\ref{theorem2} is contained in Appendix~\ref{thm2proof}. Therefore, we propose the following new objective in the outcome weighted learning: 
\begin{equation}
\label{owl3}
\begin{aligned}
& \underset{f\in \mathcal{F}}{\text{minimize}}
& & \sum_{i=1}^{n}\frac{|Y_{i}-\hat{\mu}(X_i)|}{\pi_{A_{i}}(X_{i})}\phi\left(A_{i}\times \text{sign}(Y_i-\hat{\mu}(X_i)\right)\times f(X_{i}))+ J(f),
\end{aligned}
\end{equation}
where $\hat{\mu}(X_i)$ is an estimate of $\mu(X_i)$. It is easy to see that when the residual $Y_i-\hat{\mu}(X_i)$ is positive, minimizing the objective function in $(\ref{owl3})$ encourages that the estimated ITR at $X_i$ is equal to the assigned true treatment $A_i$. Similarly, when the residual is negative, the estimated ITR tends to move away from $A_i$.

In order to use $(\ref{owl3})$, we need to plug in an estimate $\hat{\mu}(X_i)$. There are various ways to estimate the common effect $\mu(x)$. We propose a method based on the following Theorem. 
\begin{theorem}
\label{theorem3}
If $Y=\mu(X)+\delta(X)\times A+\epsilon$, then we have
\begin{equation*}
\mu(X)=E\left\{\frac{Y}{2\pi_{A}(X)}\right\}. 
\end{equation*}
Furthermore, $\mu$ satisfies 
\begin{equation*}
\mu=
\begin{aligned}
& \underset{g}{\text{argmin}}
& & E\left\{\frac{1}{\pi_{A}(X)}(Y-g(X))^2\right\}. 
\end{aligned}
\end{equation*}
\end{theorem}
The proof of Theorem~\ref{theorem3} is contained in Appendix~\ref{theorem3proof}. Similar to \cite{zhou2017}, we consider two models of $\mu(X)$. In the first one, we assume a linear model $\mu(X)=\alpha_0+\alpha^TX$. Then based on Theorem~\ref{theorem3}, we estimate regression coefficients $\alpha_0$ and $\alpha$ via the weighted least squares 
\begin{equation}
\label{commoneffect1}
\begin{aligned}
& \underset{\alpha_{0}, ~\alpha}{\text{minimize}}
& & \sum_{i=1}^{n}\frac{1}{\pi_{A_i}(X_i)}(Y_i-\alpha_{0}-\alpha^TX_i)^2,
\end{aligned}
\end{equation}
or the regularized weighted least squares 
\begin{equation}
\label{commoneffect2}
\begin{aligned}
& \underset{\alpha_{0}, ~\alpha}{\text{minimize}}
& & \sum_{i=1}^{n}\frac{1}{\pi_{A_i}(X_i)}(Y_i-\alpha_{0}-\alpha^TX_i)^2+\lambda\|\alpha\|_{1}. 
\end{aligned}
\end{equation}
In the second model, we assume a null model $\mu(X)=\alpha_0$. Then from Theorem~\ref{theorem3}, a simple estimator of $\alpha_0$ is 
\begin{equation}
\label{commoneffect3}
\hat{\alpha}_{0}=\frac{\sum_{i=1}^{n}\frac{Y_i}{\pi_{A_i}(X_i)}}{\sum_{i=1}^{n}\frac{1}{\pi_{A_i}(X_i)}}. 
\end{equation}

Finally, we consider to optimize $(\ref{owl3})$ via the boosting technique. We assume the minimizer $f^{**}_{\phi}$ in $(\ref{minimizer})$ can be characterized as additive trees. Then $(\ref{owl3})$ becomes 
\begin{equation}
\label{owl4}
\begin{aligned}
& \underset{b_{k},~1\le k\le K}{\text{minimize}}
& & \sum_{i=1}^{n}\frac{|Y_{i}-\hat{\mu}(X_i)|}{\pi_{A_{i}}(X_{i})}\phi\left(A_{i}\times \text{sign}\left(Y_i-\hat{\mu}(X_i)\right)\times \left(\sum_{k=1}^{K}b_{k}(X_i)\right) \right)+ \sum_{k=1}^{K}J(b_{k}). 
\end{aligned}
\end{equation}
Direct minimization of $(\ref{owl4})$ via the forward stagewise algorithm is challenging due to the non-quadratic form of the deviance loss. Instead, following the XGBoost algorithm, we consider its second-order approximation. More specifically, let $\hat{f}^{(t-1)}$ be the estimated additive trees at the (t-1)-th iteration. Then at the t-th iteration, we optimize the objective
\begin{equation}
\label{2ndorder}
\begin{aligned}
& \underset{b}{\text{minimize}}
& & \sum_{i=1}^{n}\frac{|Y_i-\hat{\mu}(X_i)|}{\pi_{A_i}(X_{i})}\left\{A_{i}\times \text{sign}(Y_i-\hat{\mu}(X_i))g_{i}^{(t)}b(X_{i})+\frac{1}{2}h_{i}^{(t)}b^2(X_{i})\right\}+ J(b), 
\end{aligned}
\end{equation}
where 
\begin{equation*}
g_{i}^{(t)}=\phi^{\prime}\left(A_{i}\times \text{sign}(Y_i-\hat{\mu}(X_i))\hat{f}^{(t-1)}(X_{i})\right), 
\end{equation*}
and 
\begin{equation*}
h^{(t)}_{i}=\phi^{\prime\prime}\left(A_{i}\times \text{sign}(Y_i-\hat{\mu}(X_i))\hat{f}^{(t-1)}(X_{i})\right). 
\end{equation*}
Therefore, we use split finding algorithms in XGBoost to solve $(\ref{2ndorder})$ and then apply the shrinkage procedure as $(\ref{shrinkage})$ in the end. 

In summary, we propose the following second XGBoost-based algorithm in direct learning:
\begin{algorithm}[H]
%\SetAlgoLined
{\bf Input}: data set $\{(X_{i}, A_{i}, Y_{i})\}_{i=1}^{n}$, number of iterations $K$, shrinkage parameter $\eta$ and maximum tree depth $d$. 
 \begin{enumerate}
    \item Estimate the common effect $\mu(X_i)$ for $1\le i\le n$ by $(\ref{commoneffect1}), (\ref{commoneffect2})$ or $(\ref{commoneffect3})$. 
    \item Use the XGBoost algorithm with the weighted deviance loss in $(\ref{owl3})$ to estimate $f_{\phi}^{**}(x)$. Again we denote the estimate by $\hat{f}(x)$. 
     \item Output the estimated optimal ITR: $\widehat{\mathcal{D}}(x)=\sign\left(\hat{f}\left(x\right)\right)$. 
 \end{enumerate}
 \caption{Tree boosting algorithm II in direct learning}
\end{algorithm}

%%%%%%%%%%%%%%%%%%%%%%%%%%%%%%%%%%%%%%%%%%%%%%%%%%%%%%%%%%%%%%%%%%%%%%%%%%%%%%%%

\section{Simulation studies}
\label{simusec}

In this section, we report results of simulation studies that were used to examine the performance of our algorithms introduced in Section~\ref{methodsec}. We will refer to the three proposed algorithms as \texttt{IndirectBoosting}, \texttt{DirectBoosting-I}, and \texttt{DirectBoosting-II}, respectively. 

\subsection{Simulation settings}
We generated the outcome from Model $(\ref{responserep})$. That is, for $1\le i\le n$, let 
\begin{equation*}
Y_i=\mu(X_i)+\delta(X_i)\times A_i+\epsilon_i, 
\end{equation*}
where each component of $X_i\in \mathbb{R}^{p}$ was independently generated from the uniform distribution $U(-1, 1)$, the treatment assignment $A_i$ was generated from $\mathcal{A}=\{-1, +1\}$ independently of $X_i$ with $P(A_i=-1)=P(A_i=+1)=\frac{1}{2}$, and $\epsilon_i$ was generated from the standard normal distribution $N(0, 1)$. We set $\mu(X)=1+2X_{(1)}+X_{(2)}+\frac{1}{2}X_{(3)}$, where $X_{(1)}, X_{(2)}$ and $X_{(3)}$ are the first, second, and third components of $X$. Furthermore, we considered the following five scenarios
of $\delta(X)$: 
\begin{itemize}
\item[(1)] 
\begin{equation*}
\delta(X)=3I\left(X_{(1)}\le \frac{1}{2}\right)\left(I\left(X_{(2)}>-\frac{1}{2}\right)-1\right)+1;
\end{equation*}
\item[(2)]
\begin{equation*}
\delta(X)=1.3\times \left(X_{(2)}-2X_{(1)}^2+0.3\right); 
\end{equation*}

\item[(3)] 
\begin{equation*}
\delta(X)=0.2+X_{(1)}^2+X_{(2)}^2-X_{(3)}^2-X_{(4)}^2;
\end{equation*}

\item[(4)]
\begin{equation*}
\delta(X)=3.8\left(0.8-X_{(1)}^2-X_{(2)}^2\right);
\end{equation*}

\item[(5)]
\begin{equation*}
\delta(X)=1-X_{(1)}^3+\exp{\left(X_{(3)}^2+X_{(5)}\right)}+0.6X_{(6)}-\left(X_{(7)}+X_{(8)}\right)^2.
\end{equation*}
\end{itemize}
These interaction functions are similar to those in \cite{zhao2012}, \cite{zhou2017}, and \cite{qi2019}. The first four scenarios correspond to tree-type, parabola-type, polynomial-type, and circle-type decision rules, while the last scenario was designed for simulating highly nonlinear and irregular decision rule.  Explicit formulas of the optimal ITRs in these simulation settings are provided in Appendix~\ref{simuITR}.

\subsection{Performance measures}
\label{criteria-section}

Following \cite{zhou2017} and \cite{qi2019}, we assessed the performance of the estimated optimal ITR via the estimated value function and the misclassification error rate, respectively. In particular, we use the following estimator of value function proposed by \cite{murphy2005experimental}: 
\begin{equation*}
\hat{V}(\mathcal{D})=\frac{\frac{1}{n}\sum_{i=1}^{n}\frac{Y_{i}}{\pi_{A_i}(X_i)}I(\mathcal{D}(X_i)=A_i)}{\frac{1}{n}\sum_{i=1}^{n}\frac{I(\mathcal{D}(X_i)=A_i)}{\pi_{A_i}(X_i)}},
\end{equation*}
where $\mathcal{D}$ is a given ITR and $\{(X_{i}, A_{i}, Y_{i})\}_{i=1}^{n}$ is an independent testing data. Next, let $\widehat{\mathcal{D}}$ be an estimated optimal ITR. Then the misclassification error rate is defined by 
\begin{equation*}
e=\frac{1}{n}\sum_{i=1}^{n}I\left(\mathcal{D}^*\left(X_i\right)\ne \hat{\mathcal{D}}\left(X_i\right)\right),
\end{equation*}
where $\mathcal{D}^*$ is the \emph{known} true optimal ITR in the simulation, and both of $\mathcal{D}^*$ and $\widehat{\mathcal{D}}$ are applied to the testing data. 

In our simulation studies, a testing data with 3000 observations was simulated from the joint distribution of $(X, A, Y)$ to evaluate the performance. We prefer methods with larger value function and smaller misclassification error rate.

\subsection{Implementation and simulation results}
\label{othermethods}

We compared our proposed algorithms with the following methods: 
\begin{enumerate}
\item[(1)] Q-learning described in $(\ref{indir1})$ and $(\ref{indir2})$; 
\item[(2)] $\ell_{1}$-PLS proposed by \cite{qian2011} (i.e., $(\ref{indir3})$); 
\item[(3)] D-learning proposed by \cite{qi2019} (i.e., $(\ref{opt1})$ and $(\ref{opt2})$);
\item[(4)] Linear outcome weighted learning proposed by \cite{zhao2012} (i.e., $(\ref{opt3})$);
\item[(5)] Nonlinear outcome weighted learning with the Gaussian RBF kernel defined by $K(x, x^\prime)=\exp{\left(-\frac{\|x-x^{\prime}\|_{2}^2}{2h^2}\right)}$. 
\end{enumerate}
These five methods are referred to as \texttt{Q-learning}, $\ell_{1}$-\texttt{PLS}, \texttt{D-learning}, \texttt{OWL-Linear}, and \texttt{OWL-RBF}. 

In \texttt{DirectBoosting-II}, we assumed a linear model for $\mu(X)$ and used the weighted least squares $(\ref{commoneffect1})$ to estimate the common effect. In \texttt{D-learning}, when the dimension of covariates was less than or equal to 10, we optimized the objective $(\ref{opt1})$. When the dimension of covariates was larger than 10, we estimated the decision rule via the $\ell_{1}$-regularized minimization problem $(\ref{opt2})$. Furthermore, we tuned the number of boosting iterations $K$, shrinkage parameter $\eta$, and maximum tree depth $d$ in \texttt{IndirectBoosting}, \texttt{DirectBoosting-I}, and \texttt{DirectBoosting-II} from a prespecified set based on the 10-fold cross validation, which maximized the average of the estimated value function on the validation data set. The tuning parameter $\lambda$ in other methods was selected with a same procedure.  The bandwidth $h$ in \texttt{OWL-RBF} was selected as the median of the pairwise Euclidean distance of the simulated covariates based on \cite{wu2007robust}. We implemented \texttt{IndirectBoosting}, \texttt{DirectBoosting-I}, and \texttt{DirectBoosting-II} with the XGBoost R package \cite{xgboostpackage}, and implemented \texttt{OWL-Linear} and \texttt{OWL-RBF} with the DynTxRegime R package \cite{dyntxregime}. 

For each scenario, we considered two dimensions of covariates: $p=10$ and $p=50$, and three sample sizes for training data: $n=100, n=400$ and $n=800$. We repeated the simulation 100 times. Table~\ref{p10results} and Table~\ref{p50results} show the simulation results for $p=10$ and $p=50$, respectively. We now summarize the salient points below: 

\begin{itemize}
\item  \texttt{IndirectBoosting}, \texttt{Q-learning}, and $\ell_{1}$-\texttt{PLS} are in the framework of indirect learning. \texttt{Q-learning} and $\ell_{1}$-\texttt{PLS} had similar performance in terms of both misclassification error rate and empirical value function. Our proposed method \texttt{IndirectBoosting} outperformed these two methods in Scenario 1, Scenario 2, Scenario 3, and Scenario 4 across all sampling schemes. In Scenario 5, when $(n, p)=(100, 10)$ and $(n, p)=(100, 50)$, $\ell_{1}$-\texttt{PLS} performed slightly better than \texttt{IndirectBoosting}.   

\item \texttt{D-learning} was used to compare with our proposed method \texttt{DirectBoosting-I}, both of which are in the same framework of direct learning. $\texttt{DirectBoosting-I}$ performed best in terms of both misclassification error rate and empirical value function, across all sampling schemes in all scenarios.

\item \texttt{DirectBoosting-II} has the same framework with \texttt{OWL-Linear} and \texttt{OWL-RBF}. In \cite{zhao2012}, \texttt{OWL-RBF} is designed to estimate the nonlinear optimal ITR. However, interestingly, in some scenarios (e.g., Scenario 2 and $p=10$), \texttt{OWL-RBF} failed since the misclassification error was increasing and the empirical value function was decreasing as the sample size increased from 100 to 800. Furthermore, when $p=50$, \texttt{OWL-RBF} performed poorly and was very close to make a random guess in term of misclassification error rate, across all sampling schemes in all scenarios. Clearly, our proposed method \texttt{DirectBoosting-II} beated the other two significantly in terms of both misclassification error rate and empirical value function. 
 
\item Overall, our proposed methods had much better performance compared with other five methods listed in this section.  Among three proposed methods, \texttt{IndirectBoosting} and \texttt{DirectBoosting-II} outperformed \texttt{DirectBoosting-I} in all simulation settings.  
 
\end{itemize}

\begin{table}
\caption {Sample averages (standard deviations) of misclassification error rates and empirical value functions evaluated on the testing data for five scenarios with 10 covariates. The minimal misclassification error rate and the best value function for each scenario and each sampling scheme are in bold. } 
\label{p10results}  
%\centering
\resizebox{\textwidth}{!}{
\begin{tabular}{llllllllll}
\hline \hline 
   & \multicolumn{2}{c}{$(n, p)=(100, 10)$} &  & \multicolumn{2}{c}{$(n, p)=(400, 10)$} &   & \multicolumn{2}{c}{$(n, p)=(800, 10)$} \\ 
\cline{2-3} \cline{5-6} \cline{8-9}
 & Misclassification    & Value         &              & Misclassification  & Value       &        & Misclassification & Value  \\    \hline
  &   &    &    &   {\bf Scenario 1}  &   &   &    \\ \hline 
 \texttt{IndirectBoosting}   &  0.105 (0.039)  &  1.862 (0.109)        &        &  0.021 (0.009)  & 2.071 (0.020)    &   & 0.006 (0.003)    & 2.105 (0.009)                        \\
 \texttt{DirectBoosting-I}   &  0.179 (0.050)  &  1.727 (0.110)     &        &  0.030 (0.018)  & 2.053 (0.037)    &   & 0.012 (0.009)          & 2.090 (0.021)                     \\
  \texttt{DirectBoosting-II}   &  {\bf 0.077} (0.033)   & {\bf 1.944} (0.078)       &        &  {\bf 0.010} (0.007)  & {\bf 2.090} (0.020)    &    & {\bf 0.004} (0.003)   & {\bf 2.106} (0.009)                            \\
   \texttt{Q-learning}   & 0.190 (0.037)   & 1.701 (0.087)     &       &  0.143 (0.025)  & 1.841 (0.042)          &    &0.135 (0.022)     & 1.865 (0.037)                              \\
   \texttt{$\ell_{1}$-PLS}   &  0.179 (0.052)  &  1.752 (0.100)     &        &  0.143 (0.031)  &  1.851 (0.054)      &   & 0.135 (0.023)    & 1.867 (0.040)                           \\   
    \texttt{D-learning}   &  0.234 (0.055)   &  1.580 (0.135)   &        &  0.162 (0.031)  &  1.788 (0.060)       &   & 0.147 (0.023)      &      1.832 (0.042)           \\   
     \texttt{OWL-Linear}   &  0.383 (0.091)  &   1.206 (0.216)     &        &  0.293 (0.077)  &  1.440 (0.175)      &   & 0.247 (0.058)     & 1.530 (0.150)                     \\   
      \texttt{OWL-RBF}   &  0.464 (0.278)   &    0.980 (0.386)   &        &  0.330 (0.167)  & 1.171 (0.082)           &   & 0.234 (0.062)      & 1.320 (0.082)                  \\  
      \hline
        &   &    &    &   {\bf Scenario 2}  &   &   &    \\ \hline 
 \texttt{IndirectBoosting}   & {\bf 0.158} (0.036)    &   {\bf 1.771} (0.053)      &        &  0.113 (0.014)  & 1.832 (0.023)     &    &   {\bf 0.069} (0.009)    &   {\bf 1.877} (0.015)                     \\
 \texttt{DirectBoosting-I}   & 0.268 (0.043)  & 1.543 (0.096)       &        &  0.142 (0.030)  &  1.778 (0.051)      &      &   0.114 (0.029)    &     1.820 (0.042)                 \\
  \texttt{DirectBoosting-II}   &  0.204 (0.042)  &   1.684 (0.081)    &        & {\bf 0.082} (0.017)  &  {\bf 1.864} (0.021)   &     &   0.081 (0.019)    &      1.865 (0.024)                    \\
   \texttt{Q-learning}   & 0.262 (0.023)  &   1.582 (0.051)       &        &  0.239 (0.008) &    1.645 (0.015)         &       &   0.238 (0.006)   &    1.650 (0.011)                      \\
   \texttt{$\ell_{1}$-PLS}   &0.240 (0.035)   &   1.628 (0.070)    &        & 0.226 (0.004) &    1.655 (0.018)     &      &  0.236 (0.005)   &    1.651 (0.011)                   \\   
    \texttt{D-learning}   &  0.295 (0.034)  &    1.489 (0.089)    &        &  0.250 (0.017) &     1.618 (0.033)     &       &0.242 (0.010)   &    1.636 (0.023)              \\   
     \texttt{OWL-Linear}   & 0.394 (0.073)   &   1.217 (0.205)      &        &  0.328 (0.057)  &  1.404 (0.147)  &      &  0.293 (0.051)  &       1.497 (0.114)                  \\   
      \texttt{OWL-RBF}   &  0.500 (0.137)   &   0.927 (0.443)      &        &  0.564 (0.098)  &  0.711 (0.305)   &     &  0.611 (0.046)   &       0.572  (0.150)                \\        
       \hline
        &   &    &    &   {\bf Scenario 3}  &   &   &    \\ \hline 
 \texttt{IndirectBoosting}   &  0.358 (0.040)   &   1.135 (0.056)      &        & {\bf 0.240} (0.019)  & {\bf 1.272} (0.032)       &          & {\bf 0.200} (0.014)    &    {\bf 1.309} (0.025)                  \\
 \texttt{DirectBoosting-I}   &  0.402 (0.044)   &   1.073 (0.066)      &        &  0.338 (0.032)  &   1.159 (0.043)       &       & 0.299 (0.019)       &   1.205 (0.034)              \\
  \texttt{DirectBoosting-II}   &  {\bf 0.338} (0.039)   &    {\bf 1.158} (0.056)     &        & 0.283 (0.021) &    1.227 (0.028)        &     & 0.208 (0.015)       &   1.297 (0.028)              \\
   \texttt{Q-learning}   &  0.456 (0.034)   & 1.002 (0.052)        &        &  0.412 (0.027) &   1.065 (0.043)       &      &0.391 (0.018)                &    1.095 (0.033)                \\
   \texttt{$\ell_{1}$-PLS}   &  0.628 (0.000) &    0.728 (0.000)      &        & 0.400 (0.073) &  1.086 (0.114)       &    &0.408 (0.089)             &     1.072 (0.140)            \\   
    \texttt{D-learning}   &  0.476 (0.034)   &   0.970 (0.052)      &        &  0.445 (0.030)  &   1.011 (0.050)           &    &0.423 (0.027)           &    1.047 (0.043)      \\   
     \texttt{OWL-Linear}   &  0.489 (0.036)  &   0.942 (0.063)      &        &  0.473 (0.047) &   0.971 (0.078)       &     &0.453 (0.055)             &   1.005 (0.088)          \\   
      \texttt{OWL-RBF}   &  0.475 (0.118)  &  0.965 (0.187)      &        &  0.433 (0.076) & 1.022 (0.130)          &    &0.387 (0.026)               &   1.098 (0.051)      \\        
        \hline
        &   &    &    &   {\bf Scenario 4}  &   &   &    \\ \hline 
 \texttt{IndirectBoosting}   & {\bf 0.184} (0.029)   & {\bf 2.138} (0.072)         &        & {\bf 0.092} (0.015)  &  {\bf 2.316} (0.027)        &     & {\bf 0.065} (0.011)     & {\bf 2.346} (0.019)                    \\
 \texttt{DirectBoosting-I}   & 0.225 (0.032)   &  1.998 (0.104)       &        &  0.145 (0.015) &   2.209 (0.041)         &    &  0.108 (0.011)     & 2.283 (0.028)                    \\
  \texttt{DirectBoosting-II}   & 0.207 (0.033)    &  2.036 (0.086)      &        & 0.103 (0.014)  &   2.291 (0.031)       &    & 0.069 (0.009)      & 2.342 (0.018)                  \\
   \texttt{Q-learning}   & 0.415 (0.034)   &  1.284 (0.141)      &        &  0.378 (0.012) &     1.441 (0.047)            &      & 0.373 (0.006)        & 1.463 (0.024)                    \\
   \texttt{$\ell_{1}$-PLS}   &  0.422 (0.084)  &  1.259 (0.346)    &        & 0.376 (0.011)  &   1.448 (0.045)         &      & 0.373 (0.004)        & 1.462  (0.019)                  \\   
    \texttt{D-learning}   & 0.431 (0.037)   &  1.221 (0.153)      &        &  0.394 (0.025) &   1.374 (0.102)          &      & 0.378 (0.013)          & 1.440  (0.054)           \\   
     \texttt{OWL-Linear}   & 0.459 (0.044)  &    1.112 (0.182)      &        & 0.446 (0.050)  & 1.162 (0.208)        &      &  0.422 (0.051)          & 1.261 (0.207)               \\   
      \texttt{OWL-RBF}   & 0.463 (0.110)    &    1.089 (0.453)      &        &  0.403 (0.059)  &  1.326 (0.248)        &     & 0.364 (0.014)           & 1.494 (0.054)              \\       
        \hline
        &   &    &    &   {\bf Scenario 5}  &   &   &    \\ \hline 
 \texttt{IndirectBoosting}   & 0.086 (0.015)   & 2.971 (0.051)         &        &  0.073 (0.005) &  {\bf 3.018} (0.014)     &          &0.063 (0.004)  & 3.046 (0.013)                         \\
 \texttt{DirectBoosting-I}   & 0.093 (0.019)  &  2.950 (0.061)       &        &  0.100 (0.014) &   2.938 (0.040)     &          & 0.075 (0.010)    & 3.013 (0.028)                        \\
  \texttt{DirectBoosting-II}   &0.101 (0.025)   & 2.906 (0.100)       &        &  {\bf 0.072} (0.010) &  3.015 (0.035)    &         &  {\bf 0.056} (0.007)     & {\bf 3.056} (0.022)                         \\
   \texttt{Q-learning}   & 0.090 (0.016)   &  2.972 (0.042)       &        &  0.079 (0.007) &  3.000 (0.016)          &         & 0.077 (0.004)      & 3.005 (0.011)                            \\
   \texttt{$\ell_{1}$-PLS}   &  {\bf 0.080} (0.086)  & {\bf 2.976} (0.417)        &        &  0.371 (0.411) &     1.560 (1.997)    &      &  0.076 (0.004)      & 3.008 (0.010)                      \\   
    \texttt{D-learning}   &  0.102 (0.027)   &  2.938 (0.071)        &        & 0.084 (0.011)  &    2.987 (0.026)      &        & 0.080 (0.007)        & 2.998 (0.018)               \\   
     \texttt{OWL-Linear}   &  0.239 (0.126) &   2.302 (0.582)      &        &  0.095 (0.066)&   2.939 (0.219)      &         & 0.077 (0.032)         & 3.001 (0.094)                  \\   
      \texttt{OWL-RBF}   & 0.396 (0.389)    & 1.424 (1.902)      &        &  0.228 (0.223) &    2.195 (1.133)       &          &  0.123 (0.092)     & 2.720 (0.510)                  \\           
      \hline \hline
\end{tabular}
}
\end{table}

\begin{table}
\caption {Sample averages (standard deviations) of misclassification error rates and empirical value functions evaluated on the testing data for five scenarios with 50 covariates. The minimal misclassification error rate and the best value function for each scenario and each sampling scheme are in bold. } 
\label{p50results}
%\centering
\resizebox{\textwidth}{!}{
\begin{tabular}{llllllllll}
\hline \hline 
   & \multicolumn{2}{c}{$(n, p)=(100, 50)$} &  & \multicolumn{2}{c}{$(n, p)=(400, 50)$} &   & \multicolumn{2}{c}{$(n, p)=(800, 50)$} \\ 
\cline{2-3} \cline{5-6} \cline{8-9}
   & Misclassification  &  Value             &        & Misclassification   &  Value        &        & Misclassification &  Value   \\    \hline
  &   &    &    &   {\bf Scenario 1}  &   &   &    \\ \hline 
 \texttt{IndirectBoosting}   & {\bf 0.115} (0.049)   &   {\bf 1.828} (0.137)      &        &  0.030 (0.012)  & 2.059 (0.031)               &   & 0.014 (0.006) & 2.097 (0.016)                   \\
 \texttt{DirectBoosting-I}   &0.129 (0.072) &  1.781 (0.216)      &        & 0.038 (0.017)  &    2.045 (0.040)                 &    & 0.034 (0.014)    & 2.059 (0.029)        \\
  \texttt{DirectBoosting-II}   &  0.123 (0.052)  &    1.723 (0.177)     &        &  {\bf 0.016} (0.008) &      {\bf 2.087} (0.026)          &    & {\bf 0.010} (0.005)    & {\bf 2.103} (0.014)                  \\
   \texttt{Q-learning}   &  0.465 (0.050) &  1.007 (0.116)      &        &  0.201 (0.023)  &       1.648 (0.052)                &      & 0.166 (0.015)     & 1.737 (0.030)              \\
   \texttt{$\ell_{1}$-PLS}   & 0.217 (0.056)    &   1.618 (0.106)     &        &  0.159 (0.027) &    1.768  (0.053)           &      & 0.146 (0.026)    & 1.802 (0.053)                   \\   
    \texttt{D-learning}   &   0.268 (0.071) &  1.430 (0.159)      &        & 0.108 (0.034)  &    1.850 (0.116)                  &      &   0.086 (0.032)    & 1.921 (0.078)        \\   
     \texttt{OWL-Linear}   &  0.495 (0.052) &  0.939 (0.111)     &        & 0.410 (0.045) &    1.138 (0.100)                &     &   0.376 (0.045)     & 1.225 (0.104)           \\   
      \texttt{OWL-RBF}   &  0.494 (0.307)  & 0.946 (0.370)         &        &  0.476 (0.306) &    0.968 (0.369)          &      & 0.518 (0.306)           &  0.917 (0.369)          \\  
      \hline
        &   &    &    &   {\bf Scenario 2}  &   &   &    \\ \hline 
 \texttt{IndirectBoosting}   & {\bf 0.183} (0.046)   & {\bf 1.747} (0.075)        &        & {\bf 0.090} (0.015)   &  {\bf 1.881} (0.020)       &    & 0.102 (0.011)     & 1.871 (0.016)                   \\
 \texttt{DirectBoosting-I}   & 0.310 (0.065)   &    1.476 (0.157)     &        & 0.168 (0.029)  &   1.780 (0.046)        &     & 0.117 (0.028)    & 1.846 (0.030)                   \\
  \texttt{DirectBoosting-II}   & 0.371 (0.058)   &   1.302 (0.149)       &        & 0.107 (0.024)   &   1.868 (0.029)     &     &   {\bf 0.077} (0.019)   & {\bf 1.896} (0.015)                       \\
   \texttt{Q-learning}   & 0.465 (0.038) &   1.070 (0.105)    &        & 0.263 (0.011)   &      1.617 (0.029)               &     & 0.247 (0.007)       &  1.663 (0.022)                \\
   \texttt{$\ell_{1}$-PLS}   & 0.248 (0.027)  &   1.635 (0.060)      &        &  0.233 (0.006)  &  1.700 (0.014)          &     &  0.231 (0.003)      & 1.698 (0.008)                    \\   
    \texttt{D-learning}   & 0.341 (0.049)    &  1.506 (0.130)         &        &  0.273 (0.035) &   1.664 (0.067)          &      &  0.260 (0.024)      & 1.692 (0.041)              \\   
     \texttt{OWL-Linear}   & 0.482 (0.037)   &  1.018 (0.110)        &        &  0.420 (0.033) &  1.198 (0.094)           &    &  0.380 (0.033)        & 1.302 (0.093)              \\   
      \texttt{OWL-RBF}   & 0.468 (0.141)     & 1.073 (0.511)         &        & 0.463 (0.139)   &    1.094 (0.506)        &    &   0.491 (0.144)       & 0.990 (0.523)                \\        
       \hline
        &   &    &    &   {\bf Scenario 3}  &   &   &    \\ \hline 
 \texttt{IndirectBoosting}   &  {\bf 0.385} (0.039)   &   {\bf 1.135} (0.070)      &        & {\bf 0.269} (0.020)    & {\bf 1.291} (0.032)       &    &  0.210 (0.015)     & {\bf 1.361} (0.024)                       \\
 \texttt{DirectBoosting-I}   & 0.452 (0.036)    &  1.034 (0.056)      &        &  0.381 (0.032)  &  1.138 (0.052)      &    &  0.299 (0.032)         & 1.236 (0.049)                 \\
  \texttt{DirectBoosting-II}   & 0.431 (0.037) &  1.058 (0.055)        &        &   0.300 (0.024) &  1.236 (0.041)      &    & {\bf 0.201} (0.021)         & 1.345 (0.028)                  \\
   \texttt{Q-learning}   &  0.497 (0.019)  &  0.977 (0.032)      &        &  0.456 (0.016) &     1.018 (0.032)              &   &  0.438 (0.014)       & 1.036 (0.031)                  \\
   \texttt{$\ell_{1}$-PLS}   & 0.500 (0.069)   &   0.968 (0.084)      &        &  0.415 (0.029) &   1.069 (0.041)        &    &  0.387 (0.016)       & 1.097 (0.025)                      \\   
    \texttt{D-learning}   & 0.464 (0.048) &   1.009 (0.060)      &        & 0.396 (0.036)  &    1.088 (0.043)             &     &  0.372  (0.000)        & 1.114 (0.000)          \\   
     \texttt{OWL-Linear}   &  0.502 (0.018)   &   0.969 (0.039)       &      &   0.490 (0.017) &    0.982 (0.038)     &     &    0.488 (0.020)         & 0.982 (0.041)              \\   
      \texttt{OWL-RBF}   &  0.513 (0.128)   &    0.956 (0.143)   &        & 0.482 (0.127)  &    0.990 (0.143)          &     & 0.505 (0.129)           & 0.965 (0.144)           \\        
        \hline
        &   &    &    &   {\bf Scenario 4}  &   &   &    \\ \hline 
 \texttt{IndirectBoosting}   & {\bf 0.255} (0.034)   & {\bf 1.948} (0.113)        &        & {\bf 0.134} (0.016)  & {\bf 2.266} (0.037)         &     & {\bf 0.087} (0.010) & {\bf 2.341} (0.019)                        \\
 \texttt{DirectBoosting-I}   &0.296 (0.050)  &   1.765 (0.196)     &        & 0.179 (0.017)  &    2.148 (0.047)         &     &  0.141 (0.015)   & 2.239 (0.042)                     \\
  \texttt{DirectBoosting-II}   &  0.338 (0.056)  & 1.607 (0.222)      &        &0.142 (0.016)   &  2.239 (0.044)        &     & 0.104 (0.011)     & 2.307 (0.028)                     \\
   \texttt{Q-learning}   & 0.494 (0.019) &0.985 (0.073)       &      &  0.435 (0.015) &     1.208 (0.061)                &    &  0.411 (0.014)      & 1.297 (0.056)                 \\
   \texttt{$\ell_{1}$-PLS}   &  0.454 (0.032)  &  1.133 (0.123)       &      & 0.376 (0.009) &  1.428 (0.039)            &    & 0.375 (0.025)      & 1.433 (0.093)                   \\   
    \texttt{D-learning}   &0.380 (0.044)  &  1.414 (0.159)      &        & 0.406 (0.020) &      1.318 (0.076)             &     &  0.373 (0.000)      & 1.442 (0.000)         \\   
     \texttt{OWL-Linear}   & 0.499 (0.019) &   0.962 (0.071)       &        &0.483 (0.019)    &  1.028 (0.077)           &    &  0.472 (0.020)     & 1.061 (0.083)                 \\   
      \texttt{OWL-RBF}   & 0.510 (0.128)  &   0.940 (0.466)     &        &0.515 (0.127)   &   0.921 (0.464)             &    &  0.505 (0.128)       & 0.959 (0.467)        \\       
        \hline
        &   &    &    &   {\bf Scenario 5}  &   &   &    \\ \hline 
 \texttt{IndirectBoosting}   &  0.091 (0.019)   & 2.915 (0.0626)        &        &  {\bf 0.073} (0.003)  & {\bf 2.978} (0.008)        &     & 0.066 (0.003)    & {\bf 3.001} (0.009)                    \\
 \texttt{DirectBoosting-I}   & 0.102 (0.020)  &    2.882 (0.0645)     &        & 0.077 (0.008) &  2.970 (0.018)             &    & 0.080 (0.008)    & 2.954 (0.022)                 \\
  \texttt{DirectBoosting-II}   & 0.146 (0.038)  & 2.670 (0.154)        &        &  0.075 (0.005) &   2.966 (0.021)        &      & {\bf 0.063} (0.006)    & 2.996 (0.015)                  \\
   \texttt{Q-learning}   &  0.414 (0.070)  & 1.372 (0.374)       &        &0.096 (0.010)  &       2.922 (0.030)          &         & 0.084 (0.005)     & 2.955 (0.014)                      \\
   \texttt{$\ell_{1}$-PLS}   & {\bf 0.071} (0.000)   &  {\bf 2.983} (0.000)      &        & 0.080 (0.007)   &   2.962 (0.017)     &         &  0.071 (0.000)       & 2.983  (0.000)                        \\   
    \texttt{D-learning}   & 0.124 (0.030)   & 2.815 (0.098)      &        & 0.093 (0.012)  &    2.926 (0.035)              &        & 0.086 (0.008)     & 2.950 (0.020)          \\   
     \texttt{OWL-Linear}   &0.467 (0.054)    & 1.104 (0.275)        &        & 0.307 (0.054)    &    1.966 (0.260)    &           & 0.175 (0.084)     & 2.574 (0.339)                       \\   
      \texttt{OWL-RBF}   &  0.380 (0.414)    & 1.524 (1.955)      &        & 0.491 (0.431)  &    0.998 (2.036)        &         & 0.509 (0.431)      & 0.917 (2.036)                 \\           
      \hline \hline
\end{tabular}
}
\end{table}

%%%%%%%%%%%%%%%%%%%%%%%%%%%%%%%%%%%%%%%%%%%%%%%

\section{Real data analysis}
\label{realdatasec}

We analyzed a diabetes data set which was collected from a randomized, double-blind, parallel-group Phase III trial \cite{diabetes}. This data set has been analyzed in several other papers on estimating the optimal ITR \cite{fu2014}. The randomized clinical trial was designed to compare drug efficacy of gliclazide and pioglitazone. In the comparison study, There were 1270 patients with Type 2 diabetes. Each of them was randomized to receive either pioglitazone up to 45 mg once daily or gliclazide up to 160 mg two times daily with equal probabilities. Therefore we set $\pi_{A}(X)=1/2$. The primary efficacy endpoint in the study was change from baseline $\text{HbA}_{\text{1c}}$ to the last available post-treatment value during 52 weeks. 

In our analysis, after data preprocessing, we considered 1247 patients. 624 patients received gliclazide and 623 patients received pioglitazone.  We focused on 21 baseline clinical covariates, including HDL, LDL, cholesterol, triglycerides, creatinine, fasting insulin, ALT, AST, GGT, duration of diabetes, age, weight, BMI, waist, fasting blood glucose, HomaS, HomaIR, HomaB, diastolic blood pressure, systolic blood pressure, and pulse. Detailed descriptions of these variables are in \cite{diabetes}. We compared eight methods listed in Section~\ref{othermethods}: \texttt{IndirectBoosting}, \texttt{DirectBoosting-I}, \texttt{DirectBoosting-II}, \texttt{Q-learning}, $\ell_{1}$-\texttt{PLS}, \texttt{D-learning}, \texttt{OWL-Linear}, and \texttt{OWL-RBF}. We applied the same procedures as those in Section~\ref{othermethods} to tune parameters and train different models. Finally, to evaluate performance of estimated optimal treatment rules, we performed a 10-fold cross validation. More specifically, we split the data into 10 subsets with roughly equal sizes. We estimated the optimal ITR using 9 subsets of data, and then predicted the optimal treatments for the remaining subset of patients. We repeated the procedure 10 times to obtain the predicted optimal treatment for each patient.  Following \cite{zhou2017}, we considered two performance measures. The first one was the empirical value function evaluated on each fold. The second one was p-value described below. We divided all patients into two groups, Group 1 consisted of patients whose assigned treatments were same with the estimated optimal ITRs and Group 2 consisted of the remaining patients. Let $\mu_{1}$ be the average reduction of $\text{HbA}_{\text{1c}}$ for Group 1 in the study. Let $\mu_{2}$ be the average reduction of $\text{HbA}_{\text{1c}}$ for Group 2 in the study. Then we considered the following test: 
\begin{equation*}
H_{0}: \mu_{1}=\mu_{2}, \quad\mathrm{vs.}\quad H_{A}: \mu_{1}>\mu_{2}. 
\end{equation*}
We tested the above hypotheses with Welch's t-test.  We repeated the simulation 100 times. Therefore, for each method, we obtained 1000 empirical value functions and 100 p-values. 

In our analysis, a significant p-value is less than 0.05. Table~\ref{realdataresults} reports the sample averages (standard deviations) of empirical value functions, proportions of significant p-values and medians of p-values.  In summary,  the proposed algorithms achieved larger $\text{HbA}_{\text{1c}}$ reduction compared with their corresponding competitors. The significant tests also confirmed their superior performance. 

\begin{table}[t]
\caption {Sample averages (standard deviations) of empirical value functions evaluated on each fold in cross validation, proportions of significant p-values which are less than 0.05, and medians of p-values in 100 simulations. The best value function and the largest proportion of significant p-values are in bold. } 
\label{realdataresults}
\resizebox{\textwidth}{!}{
\begin{tabular}{lccc}
\hline \hline 
   &  Value            &   Proportion of significant p-values  & Median of p-values  \\    \hline
 \texttt{IndirectBoosting}  & 1.447 (0.158)   &  {\bf 0.71}    &      0.022                  \\
 \texttt{DirectBoosting-I}   &  1.422(0.165) &  0.37      &   0.082       \\
  \texttt{DirectBoosting-II}   &  {\bf 1.448} (0.165)  &    0.69     &     0.022                  \\
   \texttt{Q-learning}   &  1.369 (0.162) &  0   &             0.500        \\
   \texttt{$\ell_{1}$-PLS}   &  1.428 (0.161)    &   0.44   & 0.060                \\   
    \texttt{D-learning}   &   1.416 (0.164) &  0.29    &     0.095         \\   
     \texttt{OWL-Linear}   &  1.360 (0.155) &  0    &    0.637            \\   
      \texttt{OWL-RBF}   &  1.363 (0.177)  & 0.04        &     0.584           \\       
      \hline \hline
\end{tabular}
}
\end{table}

%%%%%%%%%%%%%%%%%%%%%%%%%%%%%%%%%%%%%%

\section{Discussion}
\label{dis-sec}

In this paper, we have proposed three tree boosting algorithms for estimating optimal individualized treatment rules. The goal of these boosting algorithms is to improve the finite sample performance of current existing methods in both of indirect learning and direct learning. Compared with other methods discussed in the paper, the proposed algorithms achieve higher value function and lower misclassification error rate as shown in the simulation studies and real data analysis. Our tree-based methods are nonparametric in the sense that they don't assume any parametric form of decision rules. This flexibility makes the proposed algorithms extremely useful in the era of big data. On the other hand, compared with the single tree, additive trees used in these algorithms are helpful to boost the performance of estimated optimal ITRs.  

Our paper naturally suggests several venues for future work. First, \cite{baron2013reporting} points out $17.6\%$ of published randomized controlled trial in 2009 had multiple arms. Therefore it would be interesting and worthwhile to generalize our study to the multinary case. There are several recent developments in this area \cite{zhang2017, qi2019}. Here we take \cite{zhang2017} as an example. Assume the treatment space is $\mathcal{A}=\{1, 2,..., M  \}$ where $M>2$. Let 
\begin{equation*}
W_{j}=
\begin{cases}
(M-1)^{-1/2}\mathbbm{1}_{M-1} & j=1 \\
-\frac{1+M^{1/2}}{(M-1)^{3/2}}\mathbbm{1}_{M-1}+(\frac{M}{M-1})^{1/2}e_{j-1} & 2\le j\le M, 
\end{cases}
\end{equation*}
where $\mathbbm{1}_{M-1}$ is a vector of 1's with length $M-1$ and $e_{j}\in \mathbb{R}^{M-1}$ is a vector with the j-th component 1 and 0 elsewhere. $W_{j}$ can be viewed as a class label 
that identifies treatment $j$. \cite{zhang2017} proposes the following angle-based approach to estimate the decision rule: 
\begin{equation}
\label{multiopt}
\hat{f}=
\begin{aligned}
& \underset{f}{\text{argmin}}
& & \frac{1}{n}\sum_{i=1}^{n}\frac{|Y_{i}|}{\pi_{A_{i}}(X_{i})}\ell_{Y_{i}}(f(X_{i})^TW_{y_{i}})+ J(f), 
\end{aligned}
\end{equation}
where $f$ is a function from $\mathcal{X}$ to $\mathbb{R}^{k-1}$, $\ell_{Y_i}(u)=\ell(u)$ if $Y_i>0$ and $\ell_{Y_i}(u)=\ell(-u)$ if $Y_i<0$ for a convex and strictly decreasing loss function $\ell(\cdot)$. In particular, \cite{zhang2017} considers the parametric form of $f$ and uses the BFGS algorithm to solve $(\ref{multiopt})$. On the other hand, \cite{saberian2011multiclass} introduces a similar framework of multiclass boosting with coordinate descent or gradient descent. Therefore, it would be of interest to combine these two methods to develop new boosting algorithms for estimating optimal ITRs in the multinary setting.  

Second, results from boosting algorithms are not straightforward to explain. Recently, there is an active research line called interpretable machine learning, which concerns to  
make black box models explainable \cite{murdoch2019interpretable}. SHAP \cite{shap2017}, which originates from the Shapley value in cooperative game theory, is a unified and rigorous approach to explain the prediction of a machine learning model via the feature importance score. Therefore, it would be interesting to integrate SHAP to our proposed boosting algorithms. This would allow practitioners in the area of precision medicine to conduct valid personalized intervention. Third, our current work focuses on the continuous outcome. It would also be interesting to extend our methods to other types of outcome. In practice, other settings of interest in clinical studies involve the binary outcome or the survival outcome. \cite{qi2019} proposes the general direct learning framework to deal with these types, and it is quite straightforward to develop similar boosting algorithms under that framework via using appropriate loss functions. 

Finally, our current work also remains several other challenging questions. An interesting and hard question to address is exploring the theoretical relationship between the performance of our boosting algorithms and the number of iterations, which would provide insights to practical aspects of boosting. Another direction to pursue would be studying the finite sample performance of our tree boosting algorithm II in direct learning using other appropriate loss functions rather than the deviance loss. This is both of theoretical interest that has largely remained open in the robust statistics literature, and of practical interest in observational studies where the observed data set may contain contaminations.

%%%%%%%%%%%%%%%%%%%%%%%%%%%%%%%%%%%%%%%%%%%%%%%%%%%%%%

\section*{Acknowledgments}
Part of this work was completed while DW was interning at the Machine Learning and Artificial Intelligence group at Eli Lilly and Company in Indianapolis, IN. 

%%%%%%%%%%%%%%%%%%%%%%%%%%%%%%%%%%%%%%%%%%%%%%%%%%%%%%

\nocite{*}
\bibliographystyle{plain}
\bibliography{refs}

\appendix

\section{Proofs}
In this Appendix, we provide poofs of theorems and propositions in the paper.

\subsection{Proof of Theorem~\ref{fisherconsistency-binary}}
\label{fisherconsistency-binary-proof}

\begin{proof}

For $x\in \mathcal{X}$, let $S_{\phi}(w)=E\left\{Y\frac{\phi(Aw)}{\pi_{A}(X)}|X=x\right\}$. Then we have 
\begin{equation*}
\begin{aligned}
&f^*_{\phi}(x)= \underset{w}{\text{argmin}}
& &  S_{\phi}(w). 
\end{aligned}
\end{equation*}
Furthermore, we have
\begin{equation*}
S_{\phi}(w)=E\left(Y|X=x, A=1\right)\phi(w)+E\left(Y|X=x, A=-1\right)\phi(-w). 
\end{equation*}
Therefore, 
\begin{equation*}
\frac{d S_{\phi}(w)}{dw}|_{w=0}=\left\{E(Y|X=x, A=1)-E(Y|X=x, A=-1)\right\}\phi^{\prime}(0). 
\end{equation*}
When $\mathcal{D}^*(x)=+1$, we have $E(Y|X=x, A=1)>E(Y|X=x, A=-1)$. Thus, when $\phi^{\prime}(0)<0$, we have $\frac{d S_{\phi}(w)}{dw}|_{w=0}<0$. Therefore, there exists a constant $\delta>0$, such that
\begin{equation*}
S_{\phi}(\delta)\le S_{\phi}(0)+S_{\phi}^{\prime}(0)\frac{\delta}{2}<S_{\phi}(0). 
\end{equation*}

On the other hand, since we assume $Y$ is positive, so $E(Y|X=x, A=1)$ and $E(Y|X=x, A=-1)$ are both positive. Therefore when $\phi$ is convex, $S_{\phi}(w)$ is also convex. Hence, for $w<\frac{\delta}{4}$, 
\begin{equation*}
S_{\phi}(w)\ge S_{\phi}(0)+S_{\phi}^{\prime}(0)w>S_{\phi}(0)+S_{\phi}^{\prime}(0)\frac{\delta}{4}>S_{\phi}(0)+S^{\prime}_{\phi}(0)\frac{\delta}{2}\ge S_{\phi}(\delta),
\end{equation*}
which implies $f^{*}_{\phi}(x)>0$. 

Similarly, when $\mathcal{D}^*(x)=-1$, we can show that $f^*_{\phi}(x)<0$. Therefore the proof is complete. 

\end{proof}

%%%%%%%%%%%%%%%%%%%%%%%%%%%%%%%%%%%%%%%%%%%%

\subsection{Proof of Theorem~\ref{theorem2}}
\label{thm2proof}

\begin{proof}
For a fixed $x$, let 
\begin{equation*}
S_{\phi}(w)=E\left\{\frac{|Y-\mu(X)|}{\pi_{A}(X)}\phi\left(wA\times\mathrm{sign}\left(Y-\mu(X)\right)\right)|X=x\right\}.
\end{equation*}
Furthermore, we have
\begin{equation*}
\begin{split}
S_{\phi}(w)=E\left\{|Y-\mu(X)|\phi(w\times \mathrm{sign}(Y-\mu(X)))|X=x, A=1\right\} \\
+E\left\{|Y-\mu(X)|\phi(-w\times \mathrm{sign}(Y-\mu(X)))|X=x, A=-1\right\} \\
=a_{1}\phi(w)+a_{2}\phi(-w), 
\end{split}
\end{equation*}
where
\begin{equation*}
a_{1}=E\left\{(Y-\mu(X))I(Y>\mu(X))|X=x, A=1\right\}+E\left\{(\mu(X)-Y)I(Y<\mu(X))|X=x, A=-1\right\}, 
\end{equation*}
and 
\begin{equation*}
a_{2}=E\left\{(Y-\mu(X))I(Y>\mu(X))|X=x, A=-1\right\}+E\left\{(\mu(X)-Y)I(Y<\mu(X))|X=x, A=1\right\}. 
\end{equation*}
Note that $a_{1}$ and $a_{2}$ are positive, and 
\begin{equation*}
\begin{split}
a_{1}-a_{2}=E\left\{Y-\mu(X)|X=x, A=1\right\}+E\left\{\mu(X)-Y|X=x, A=-1\right\} \\
=E\left\{Y|X=x, A=1\right\}-E\left\{Y|X=x, A=-1\right\}. 
\end{split}
\end{equation*}
We first consider the case that $\mathcal{D}^*(x)=1$, then $a_{1}-a_{2}>0$. Therefore we have
\begin{equation*}
S^{\prime}_{\phi}(0)= (a_{1}-a_{2})\phi^{\prime}(0)<0. 
\end{equation*}
Using a similar argument with that in Theorem~\ref{fisherconsistency-binary}, we have $f_{\phi}^{**}(x)>0$. When $\mathcal{D}^*(x)=-1$, we can also show $f^{**}_{\phi}(x)<0$. Hence the proof is complete. 

\end{proof}

%%%%%%%%%%%%%%%%%%%%%%%%%%%%%%%%%%%%%%

\subsection{Proof of Theorem~\ref{theorem3}}
\label{theorem3proof}

\begin{proof}
Note that  
\begin{equation*}
E(Y|X, A=1)=\mu(X)+\delta(X),\quad E(Y|X, A=-1)=\mu(X)-\delta(X),
\end{equation*}
so we have 
\begin{equation*}
\mu(X)=\frac{1}{2}\left\{E(Y|X, A=1)+E(Y|X, A=-1)\right\}=E\left\{\frac{Y}{2\pi_{A}(X)}|X\right\}. 
\end{equation*}
Next, for a fixed $x$, let 
\begin{equation*}
L(g)=E\left\{\frac{1}{\pi_{A}(X)}(Y-g)^2|X=x\right\}. 
\end{equation*}
Then taking derivative over $g$ and setting it to 0, we have 
\begin{equation*}
E\left\{\frac{1}{\pi_{A}(X)}(Y-g)|X=x\right\}=0.
\end{equation*}
Therefore, we have 
\begin{equation*}
g=E\left\{\frac{Y}{2\pi_{A}(X)}\right\}. 
\end{equation*}
Hence the proof is complete. 
\end{proof}

%%%%%%%%%%%%%%%%%%%%%%%%%%%%%%%%%%%%%%%%%%%%%%%%%%%%%%%%%%%%

\subsection{Proof of Proposition~\ref{equivalent-itr}}
\label{equivalent-itr-proof}

\begin{proof}
For any ITR $\mathcal{D}$, following $(\ref{valuefunction})$, we have 
\begin{equation*}
\begin{split}
V(\mathcal{D})=E\left\{Y\frac{I(A=\mathcal{D}(X))}{\pi_{A}(X)}\right\} \\
=E_{X, A}\left[E_{Y}\left\{Y\frac{I(A=\mathcal{D}(X))}{\pi_{A}(X)}  |X, A\right\}\right] \\
=E_{X, A}\left\{ \frac{I(A=\mathcal{D}(X))}{\pi_{A}(X)}Q(X, A)  \right\} \\
=E_{X}\left[E_{A}\left\{ \frac{I(A=\mathcal{D}(X))}{\pi_{A}(X)}Q(X, A) |X \right\}\right] \\
=E_{X}\left\{\sum_{a\in \mathcal{A}}I(a=\mathcal{D}(X))Q(X, a) \right\} \\
=E_{X}\left\{Q(X, \mathcal{D}(X)) \right\}. 
\end{split}
\end{equation*}
Let $d(x)=\text{argmax}_{a\in \mathcal{A}}Q(x, a)$. Then we have 
\begin{equation*}
V(\mathcal{D}^*)=E_{X}\left\{Q(X, \mathcal{D}^*(X))\right\}\le E_{X}\left\{\max_{a\in \mathcal{A}}Q(X, a)\right\}=V(d). 
\end{equation*}
On the other hand, by definition of $\mathcal{D}^*$, we have $V(\mathcal{D}^*)\ge V(d)$. Therefore, we prove the result. 
\end{proof}

%%%%%%%%%%%%%%%%%%%%%%%%%%%%%%%%%%%%%%%%%%%%%%%%%%%%%%%%%%%%%%%

\subsection{Proof of Proposition~\ref{estflemma}}
\label{estflemmaproof}
\begin{proof}
For a fixed $x$, let 
\begin{equation*}
L(g)=E\left\{\frac{1}{\pi_{A}(X)}(2YA-g)^2|X=x\right\}.
\end{equation*}
Then taking the derivative over $g$ and setting it to 0, we have 
\begin{equation*}
E\left\{\frac{1}{\pi_{A}(X)}(2YA-g)|X=x\right\}=0.
\end{equation*}
Furthermore, we have
\begin{equation*}
E\left\{\frac{1}{\pi_{A}(X)}|X=x\right\}=2. 
\end{equation*}
Hence, we have 
\begin{equation*}
g=E\left\{\frac{YA}{\pi_{A}(X)}|X=x\right\}.
\end{equation*}
\end{proof}

%%%%%%%%%%%%%%%%%%%%%%%%%%%%%%%%%%%%%%

\section{Optimal ITRs in simulation studies}
\label{simuITR}

In this Appendix, we provide the true optimal ITRs, which are known when generating the simulated data. In Scenario (1), the optimal decision rule is 
\[\mathcal{D}^{*}(X)=
 \begin{cases} 
      -1 & X_{(1)}\le \frac{1}{2}~\text{and}~X_{(2)}\le -\frac{1}{2}\\
      1 & \text{otherwise},
   \end{cases}
\]
which is a decision tree. In Scenario (2), the optimal decision rule is a parabola, which can be expressed as 
\[\mathcal{D}^{*}(X)=
 \begin{cases} 
      1 & X_{(2)}-2X_{(1)}^2+0.3>0\\
      -1 & X_{(2)}-2X_{(1)}^2+0.3<0. 
   \end{cases}
\]
The optimal decision rule in Scenario (3) is a polynomial of degree 2: 
\[\mathcal{D}^{*}(X)=
 \begin{cases} 
      1 & 0.2+X_{(1)}^2+X_{(2)}^2-X_{(3)}^2-X_{(4)}^2>0\\
      -1 & 0.2+X_{(1)}^2+X_{(2)}^2-X_{(3)}^2-X_{(4)}^2<0. 
   \end{cases}
\]
The optimal decision rule in Scenario (4) is a circle:  
\[\mathcal{D}^{*}(X)=
 \begin{cases} 
      1 & X_{(1)}^2+X_{(2)}^2<0.8\\
      -1 &  X_{(1)}^2+X_{(2)}^2>0.8. 
         \end{cases}
\]
In Scenario (5), the optimal ITR is highly nonlinear which takes the form as follows:  
\[\mathcal{D}^{*}(X)=
 \begin{cases} 
      1 & 1-X_{(1)}^3+\exp{(X_{(3)}^2+X_{(5)})}+0.6X_{(6)}-(X_{(7)}+X_{(8)})^2>0\\
      -1 &   1-X_{(1)}^3+\exp{(X_{(3)}^2+X_{(5)})}+0.6X_{(6)}-(X_{(7)}+X_{(8)})^2<0. 
         \end{cases}
\]

%%%%%%%%%%%%%%%%%%%%%%%%%%%%%%%%%%%%%%%%%%%%%%%%%%%%%%%%%%

\end{document}